\documentclass[10pt,journal,compsoc]{IEEEtran}

\ifCLASSOPTIONcompsoc

  \usepackage[nocompress]{cite}
\else

  \usepackage{cite}
\fi

\ifCLASSINFOpdf
\else
\fi

\usepackage{tabu}
\usepackage{subcaption,graphicx}
\usepackage{amsfonts}
\usepackage{amssymb}
\usepackage{amsmath}
\usepackage{threeparttable}
\usepackage{makecell}
\usepackage{multirow}
\usepackage{pifont}

\newcommand{\xmark}{\text{\ding{55}}}
\begin{document}

\title{Normalization Before Shaking Toward Learning Symmetrically Distributed Representation Without Margin in Speech Emotion Recognition}

\author{Che-Wei~Huang,~\IEEEmembership{Student~Member,~IEEE,}
        and~Shrikanth~S~Narayanan,~\IEEEmembership{Fellow,~IEEE}
\IEEEcompsocitemizethanks{\IEEEcompsocthanksitem 
The authors are with the Signal Analysis and Interpretation Laboratory (SAIL), Department of Electrical Engineering and Signal and Image Processing Institute, University of Southern California, Los Angeles, CA 90089 USA\protect\\
E-mail: cheweihu@usc.edu; shri@sipi.usc.edu
\IEEEcompsocthanksitem This work is supported by NSF.
}
}

\markboth{Submission to The IEEE Transactions}
{Che-Wei Huang \MakeLowercase{\textit{et al.}}: Submission to the IEEE Transaction}

\IEEEtitleabstractindextext{%
\begin{abstract}
Regularization is crucial to the success of many practical deep learning models, in particular in a more often than not scenario where there are only a few to a moderate number of accessible training samples.
In addition to weight decay, data augmentation and dropout, regularization based on multi-branch architectures, such as Shake-Shake regularization, has been proven successful in many applications and attracted more and more attention.
However, beyond model-based representation augmentation, it is unclear how Shake-Shake regularization helps to provide further improvement on classification tasks, let alone the baffling interaction between batch normalization and shaking.
In this work, we present our investigation on Shake-Shake regularization, drawing connections to the vicinal risk minimization principle and discriminative feature learning in verification tasks.
Furthermore, we identify a strong resemblance between batch normalized residual blocks and batch normalized recurrent neural networks, where both of them share a similar convergence behavior, which could be mitigated by a proper initialization of batch normalization.
Based on the findings, our experiments on speech emotion recognition demonstrate simultaneously an improvement on the classification accuracy and a reduction on the generalization gap both with statistical significance.

\end{abstract}

}

\maketitle

\IEEEdisplaynontitleabstractindextext

\IEEEpeerreviewmaketitle

\IEEEraisesectionheading{\section{Introduction}}
\label{sec:introduction}
\IEEEPARstart{D}eep convolutional neural networks have been successfully applied to several pattern recognition tasks such as image recognition \cite{He_2016_CVPR}, machine translation \cite{gehring2017} and speech emotion recognition \cite{Huang2017}. 
Currently, to successfully train a deep neural network, one needs either a sufficient number of training samples to implicitly regularize the learning process, or employ techniques like weight decay and dropout \cite{Srivastava:2014} and its variants to explicitly keep the model from over-fitting. 

In the recent years, one of the most popular and successful architectures is the residual neural network (ResNet) \cite{He_2016_CVPR} and its variant ResNeXt \cite{Xie2016} with multiple residual branches. 
The ResNet architecture was designed based on a key assumption that it is more efficient to optimize the residual term than the original task mapping. 
Since then, a great deal of effort in machine learning and computer vision has been dedicated to study the multi-branch architecture.
    
Deep convolutional neural networks have also gained much attention in the community of affective computing mainly because of its outstanding ability to formulate discriminative features for the top-layer classifier. 
Usually the number of parameters in a model is far more than the number of training samples and thus it requires heavy regularization to train deep neural networks for affective  computing. 
However, since the introduction of batch normalization \cite{batchnorm}, the gains obtained by using dropout for regularization have decreased \cite{batchnorm,Zagoruyko2016WRN,Huang2016}. 
A recent work dedicated to study the disharmony between dropout and batch normalization \cite{batchnorm_dropout} suggests that dropout introduces a variance shift between training and testing, which causes batch normalization to malfunction if batch normalization is placed after dropout, which severely limits the application of successful architectures such as ResNet or the application of dropout to the top-most fully connected layers.
Yet, multi-branch architectures have emerged as a promising alternative for regularizing convolutional layers.

Regularization techniques based on multi-branch architectures such as Shake-Shake \cite{gastaldi2017} and ShakeDrop \cite{shakedrop} have delivered impressive performances on standard image datasets such as the CIFAR-10 \cite{cifar10} dataset. 
In a clever way, both of them utilize multiple branches to learn different aspects of the relevant information and then a summation in the end follows for information alignment among branches. 
Also, both of Shake-Shake and ShakeDrop regularizations emphasize on the important interaction between batch normalization and shaking.
However, none of them gave an explanation for this phenomenon, other than a brief discussion on limiting the strength of shaking.
Instead of using multiple branches, a recent work \cite{randomproj} based on a mixture of experts showed that randomly projecting samples is able to break the structure of adversarial noise that could easily confound the model and as a result mislead the learning process. Despite not being an end-to-end approach, it shares the same idea of integrating multiple streams of model-based diversity.  

In addition, a recent trend of studies on data augmentation, based on the Vicinal Risk Minimization (VRM) \cite{VRM2000} principle, proposed to interpolate and/or extrapolate training samples in feature space, for example, \cite{DeVries2017} and \cite{SMOTE2002}.
Szegedy et al. \cite{Szegedy2016} studied regularization of training by label smoothing.
Furthermore, Mixup \cite{Mixup2018} performed convex combinations of pairs of feature and label to demonstrate that expanding the coverage of training samples in feature space, leaving little to none of margin between classes, could not only improve the performance of a model but also make it robust to adversarial samples.
Effectively, Mixup reduces the uncertainty in prediction of a testing sample lying outside of the coverage of training samples in feature space by linear interpolation.
Based on Mixup, Manifold Mixup \cite{ManifoldMixup2018} called for mixing intermediate representations instead of raw inputs.

Our work follows the model-based representation augmentation thread like Shake-Shake, ShakeDrop regularization and Manifold Mixup.
In this work, we study the Shake-Shake regularized ResNeXt for speech emotion recognition. 
In addition to shaking the entire spectral-temporal feature maps with the same strength, we propose to address different spectral sub-bands independently based on our hypothesis of the non-uniform distribution of affective information over the spectral axis. 
Furthermore, we investigate and come up with an explanation for the ability of shaking regularization to improve classification tasks and its crucial interaction with batch normalization.
In order to achieve our goal, we conduct ablation studies on MNIST \cite{lecun-mnisthandwrittendigit-2010} and CIFAR-10 datasets to highlight a subtle difference in the requirement of optimal embeddings by classification tasks based on the VRM principle and by verification tasks.
In addition, we identify a strong resemblance in the mathematical formulation between batch normalized residual blocks and batch normalized recurrent neural networks, where both of them suffer from a shared issue: \textit{faster convergence but more over-fitting} and could be fixed by the same technique.

Our contributions are multi-fold. 
First, our work explains with visualization the key factor to the success of shaking regularization and the crucial property that batch normalization plays in a shaking regularized architecture, drawing a connection between shaking and the VRM principle, and between batch normalization and discriminative embedding learning.
Second, to the best of our knowledge, our work is the first to highlight the subtle difference in the requirement of optimal embeddings by classification tasks and by verification tasks.
Third, our work identify the resemblance between batch normalized residual blocks and batch normalized recurrent neural networks, and the shared issue they have.
Based on the solution to batch normalized recurrent neural networks, we demonstrate a significant reduction on the generalization gap, i.e. reduced over-fitting, in a batch-normalized shaking-regularized ResNeXt for speech emotion recognition without sacrificing the validation accuracy.

The outline of this paper is as follows. 
We review related work in the next section, including Shake-Shake regularization and its variants, discriminative feature learning in verification tasks and the vicinal risk minimization principle.
Section \ref{sec:subband_shaking} introduces sub-band shaking.
Section \ref{sec:bn_shaking} presents batch normalized shaking, including ablation studies on MNIST and CIFAR-10 datasets, and the identification of batch normalized residual blocks with batch normalized recurrent neural networks.
Section \ref{sec:dataexps} and \ref{sec:exp_results} cover the datasets, the network architecture, the experimental setup and the results for speech emotion recognition.
Section \ref{sec:conclu} concludes our work.

\section{Related Work}
\label{sec:related_work}
In this section, we review some of the recently developed techniques on which we base our work and those that are related to our findings. 
The first subsection covers Shake-Shake regularization \cite{gastaldi2017} and its variants, including Sub-band Shaking \cite{Huang2018a}, ShakeDrop \cite{shakedrop} and Stochastic Shake-Shake \cite{Huang2018b} regularization.
Another subsection is devoted to end-to-end discriminative feature learning algorithms in face verification, in particular the thread of work that focus on the large-margin and symmetrical representation learning in \cite{liu2016large, Liu_2017_CVPR, CosFace2018, deng2018arcface, ccl2018}.   
In the last subsection, we talk about a recent success in supervised learning based on the Vicinal Risk Minimization (VRM) principle \cite{VRM2000}, called Mixup \cite{Mixup2018}.
Moreover, we highlight one subtle but definite difference in the desired quality of intermediate representations between classification and verification tasks.

\subsection{Shake-Shake Regularization and Its Variants}
\label{subsec:shakeshake}

\begin{figure}[t]
  \centering
  \includegraphics[width=0.45\textwidth,height=4cm]{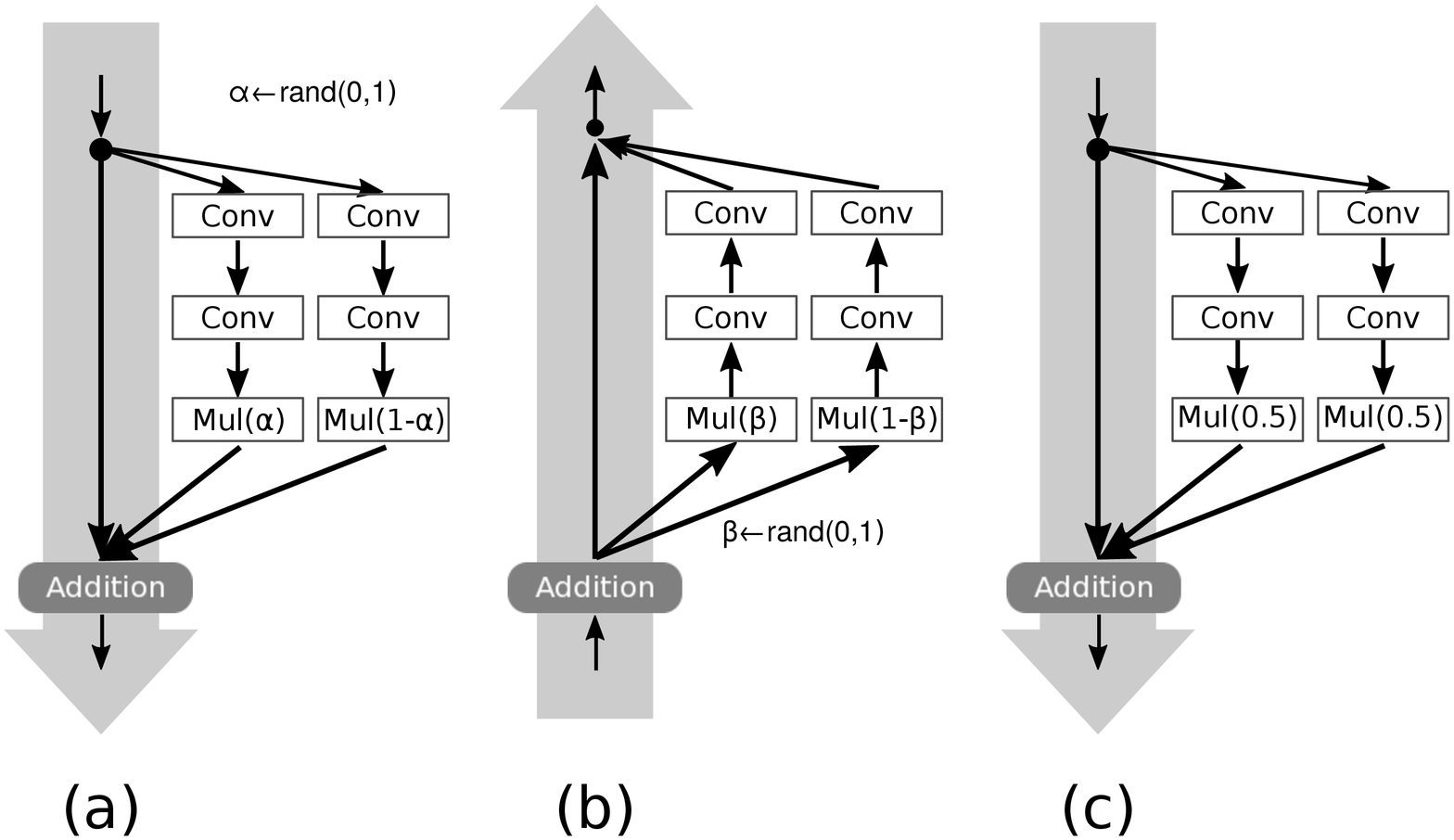}
  \caption{An overview of a $3$-branch Shake-Shake regularized residual block. 
  (a) Forward propagation during the training phase 
  (b) Backward propagation during the training phase 
  (c) Testing phase. 
  The coefficients $\alpha$ and $\beta$ are sampled from the uniform distribution over $[0,1]$ to scale down the forward and backward flows during the training phase.}
\label{fig:shakeshake}
\end{figure}

\begin{figure*}[th]
  \footnotesize
  \centering
  \includegraphics[width=0.8\linewidth, height=3in]{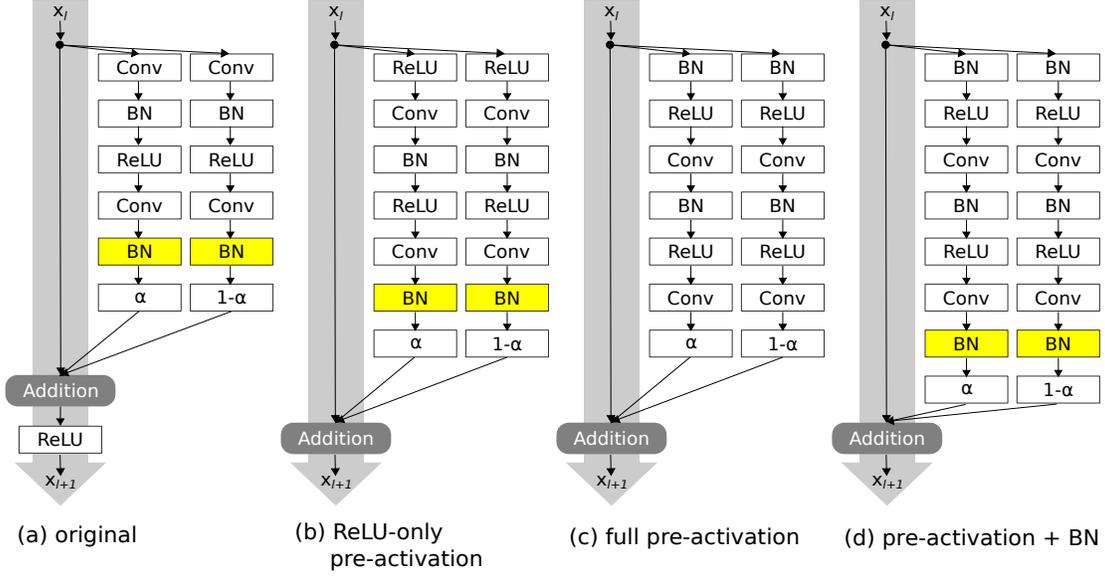}
  \caption{Shaking regularized ResNeXt architectures with different layouts introduced in \cite{He_2016_ECCV}}
  \label{fig:resnetconfig}
\end{figure*}

Shake-Shake regularization is a recently proposed technique to regularize training
 of deep convolutional neural networks for image recognition tasks.
This regularization technique based on multi-branch architectures promotes stochastic mixtures of forward and backward propagations from network branches in order to create a flow of model-based adversarial learning samples/gradients during the training phase.  
Owing to it excellent ability to combat over-fitting even in the presence of batch normalization, the Shake-Shake regularized $3$-branch residual neural network \cite{gastaldi2017} has achieved one of current state-of-the-art performances on the CIFAR-10 image dataset.

An overview of a $3$-branch Shake-Shake regularized ResNeXt is depicted in Fig. \ref{fig:shakeshake}. 
Shake-Shake regularization adds to the aggregate of the output of each branch an additional layer, called the shaking layer, to randomly generate adversarial flows in the following way:
\begin{equation}
\mathbf{\text{ResBlock}}^N(\mathbf{X}) = \mathbf{X} + \sum_{n=1}^{N} \text{Shaking}\left(\left\{\mathbf{B}_n(\mathbf{X})\right\}_{n=1}^{N}\right)
\nonumber
\end{equation}
where in the forward propagation for $\mathbf{a} = [\alpha_1, \cdots, \alpha_N]$ sampled from the ($N$$-$$1$)-simplex (Fig. \ref{fig:shakeshake} (a))
\begin{equation}
\mathbf{\text{ResBlock}}^N(\mathbf{X}) = \mathbf{X} + \sum_{n=1}^N \alpha_n \mathbf{B}_n(\mathbf{X}),
\nonumber
\end{equation}
while in the backward propagation for $\mathbf{b} = [\beta_1, \cdots, \beta_N]$ sampled from the ($N$$-$$1$)-simplex and $\mathbf{g}$ the gradient from the top layer, the gradient entering into $\mathbf{B}_n(x)$ is $\beta_n\mathbf{g}$ (Fig. \ref{fig:shakeshake} (b)).
At testing time, the expected model is then evaluated for inference by taking the expectation of the random sources in the architecture (Fig. \ref{fig:shakeshake} (c)).
In each mini-batch, to apply scaling coefficients $\alpha$ or $\beta$ either on the entire mini-batch or on each individual sample independently can also make a difference. 
 
Instead of the commonly known original or the fully pre-activation residual block, Shake-Shake regularization is proposed with the ReLU-only pre-activation residual block. 
Refer to Fig. \ref{fig:resnetconfig} for more details.
In addition, it has been shown that shaking with the absence of both batch normalization layers could cause the training process to diverge. 
One proposed remedy to this situation in \cite{gastaldi2017} is to employ a shallower architecture and more importantly to reduce the range of values $\alpha$ can take on, i.e. to reduce the strength of shaking.

ShakeDrop regularization \cite{shakedrop}, based on the same idea of model-based representation augmentation but on the deep pyramidal residual architecture, reached an improved accuracy on CIFAR-10/100 datasets.
The authors further empirically observed that each residual block should end with batch normalization before the shaking layer to prevent training from diverging. 

In our previous work on acoustic sub-band shaking \cite{Huang2018a} and stochastic Shake-Shake regularization \cite{Huang2018b} for affective computing from speech, we found that in a fully pre-activation architecture without a batch normalization layer right before shaking, the shaking mechanism contributes much more to constraining the learning process than to boosting the generalization power. 
In addition, we showed methods to relax or control the impact of shaking, either deterministically by sub-band shaking or probabilistically by randomly turning off shaking, in order to trade-off between a higher accuracy and reduced over-fitting.
Fortunately, with appropriate hyper-parameters, we could achieve both with statistical significance, compared to the baseline.

All these findings indicated there is a close interaction between shaking and batch normalization.
However, these studies did not give an explanation for the crucial location of batch normalization in a shaking regularized architecture, other than a brief discussion of the range of $\alpha$.

\subsection{End-to-End Discriminative Feature Learning}
\label{subsec:large_margin}
Recently, there has been a trend to focus on the design of loss functions so that a neural network supervised by such a loss function is able to formulate more discriminative features, usually compared to the ordinary softmax loss, for face verification. 
Inspired by the contrastive loss \cite{Contrastive2006} and the triplet loss \cite{Triplet2015}, the main design objective aims to simultaneously minimize intra-class dispersion and maximize inter-class margin.
However, using the contrastive loss or the triplets loss often involves training with pairs and triplets on the order of $O(N^2)$ and $O(N^3)$ when $N$ is the number of training samples, or one has to rely on a carefully selected sampling strategy.

A series of work hence reviewed the interpretation of the softmax loss as a normalized exponential of inner products between feature vector and class center vectors, and came up with various modifications for achieving the aforementioned design objective, including the large-margin softmax \cite{liu2016large}, SphereFace \cite{Liu_2017_CVPR}, CosFace \cite{CosFace2018}, ArcFace \cite{deng2018arcface} and Centralized Coordinate Learning (CCL) \cite{ccl2018}.

Different from the other four modifications, CCL distributes features dispersedly by centralizing the features to the origin of the space during the learning process so that feature vectors from different classes can be more separable in terms of a large angle between neighboring classes, and ideally symmetrically distributed in the whole feature space.
The CCL loss is presented as follows:
\begin{equation}
\mathcal{L} = \frac{-1}{N} \sum_i^N \log
\frac{e^{\Phi(\mathbf{x}_i)\cos(\theta_{y_i})}}
{\sum_j^K e^{\Phi(\mathbf{x}_i)\cos(\theta_{y_j})}}
\label{eq:ccl_softmax}
\end{equation}
where
\begin{equation}
\Phi(\mathbf{x}_i)_j = 
\frac{\mathbf{x}_{ij} - \mathbf{o}_j}
{\mathbf{\sigma}_j}.
\label{eq:ccl}
\end{equation}
$\Phi(\mathbf{x}_i)_j$ and $\mathbf{x}_{ij}$ are $j$-th coordinate of $\Phi(\mathbf{x}_i)$ and $\mathbf{x}_i$, respectively, and $\theta_{y_j}$ is the angle between $\mathbf{x}_i$ and the class center vector $\mathbf{w}_{y_j}$.
It is immediately clear that Eq. (\ref{eq:ccl}) resembles the famous batch normalization:
\begin{equation}
\Phi(\mathbf{x}_i)_j = 
\gamma_j
\frac{\mathbf{x}_{ij} - \mathbf{o}_j}
{\mathbf{\sigma}_j}
+\beta_j
\label{eq:bn}
\end{equation}
except that the trainable affine transformation, defined by $\gamma$ and $\beta$, after the normalization are missing in the formulation.
In Eq. (\ref{eq:ccl}) and (\ref{eq:bn}), $\sigma$ and $\mathbf{o}$ are running standard deviation and running mean updated per mini-batch during training. 
In fact, it was showed that there would be slight degradation in performance when the trainable affine transformation is employed. In the rest of this work, we use batch normalization or CCL interchangeably to refer to this discriminative feature learning. 


\subsection{Vicinal Risk Minimization and Mixup}
\label{subsec:vrmNmixup}
In statistical learning theory, because the data distribution $\mathcal{P}(x,y)$ is unknown in most practical applications, one may approximate it by the empirical distribution
\begin{equation}
  \label{eq:delta_approx_p}
  \mathcal{P}_\delta(x,y) = 
  \frac{1}{N} \sum_n \delta(x=x_n,y=y_n),
\end{equation}
where $\delta(x=x_n,y=y_n)$ is a Dirac centered at $(x_n,y_n)$, and $\mathcal{D}=\{(x_n,y_n):1\leq n\ \leq N\}$ is a given training set sampled from $\mathcal{P}(x,y)$.
This paradigm of learning via minimizing a risk based on the approximation in Eq. (\ref{eq:delta_approx_p}) is referred to as the Empirical Risk Minimization (ERM) \cite{ERM1998}.

Instead of a Dirac function, one may employ a vicinity function to take into consideration the vicinity of a true training pair $(x_n,y_n)$, and this alternative is therefore called the Vicinal Risk Minimization (VRM) principle \cite{VRM2000}.
The Gaussian vicinity \cite{VRM2000} $\nu(x,y|x_n,y_n) = \mathcal{N}(x-x_n,\sigma^2)\delta(y=y_n)$ is one of the well-known examples, which is equivalent to data augmentation with additive Gaussian noise.

Recently, a novel vicinity function, called Mixup \cite{Mixup2018}, is proposed to cover the entire convex hull of training pairs:
\begin{eqnarray}
  \label{eq:mixup}
  \nu(x,y|x_n,y_n) 
  &=& 
  \frac{1}{N} \sum_m
  \mathbb{E}_\lambda
  \left[
    \delta(
    x=u_m, y=v_m
    )
  \right],\\
  u_m
  &=&
  \lambda \cdot x_n + (1-\lambda) \cdot x_m,\\
  v_m
  &=&
  \lambda \cdot y_n + (1-\lambda) \cdot y_m,
\end{eqnarray}
where $\lambda\in[0,1]$.
This data augmentation scheme by Mixup is believed to reduce the amount of undesirable oscillations when predicting outside the original training examples as it fills in the feature space between classes by convex combinations of true training pairs, and hence contribute both to improvement on classification accuracy as well as robustness to adversarial examples.

\paragraph{\textbf{\textit{Classification versus Verification}}}
One could easily observe that the directions for improving classification tasks based on VRM and for refining embeddings in verification tasks based on the aforementioned design objective are drastically different, where the former strives to minimize or eliminate the margin between classes while the latter aims to ensure a definite minimum margin. 
We provide more details in Section \ref{sec:bn_shaking}.

\section{Sub-band Shaking}
\label{sec:subband_shaking}
Shake-Shake regularization delivers different results depending on the strength of the shaking, in terms of the range of values $\alpha$ (or $\beta$) takes on as well as whether the same $\alpha$ (or $\beta$) is shared within a mini-batch.
In addition to batch- or sample-wise shaking, when it comes to the area of acoustic processing, there is another orthogonal dimension to consider: the spectral domain. 
Leveraging domain knowledge, our first proposed models are based on a simple but plausible hypothesis that affective information is distributed non-uniformly over the spectral axis \cite{Lee04emotionrecognition}. 
Therefore, there is no reason to enforce the entire spectral axis to be shaken with the same strength concurrently. 
Furthermore, adversarial noise may exist and extend over the spectral axis. 
By deliberately shaking spectral sub-bands independently, the structure of adversarial noise may be broken and become less confounding to the model.  

There has been work on multi-stream framework in speech processing. 
For example, Mallidi et al. \cite{mallidi2016} designed a robust speech recognition system using multiple streams, each of them attending to a different part of the feature space, to fight against noise. 
However, lacking both multiple branches and the final information alignment, the design philosophy is fundamentally different from that of multi-branch architectures. 
In fact, sub-band shaking could be viewed as a bridge between the multi-stream framework and the multi-branch architecture. 

\begin{figure}[th]
  \centering
  \includegraphics[width=0.4\textwidth,height=3cm]{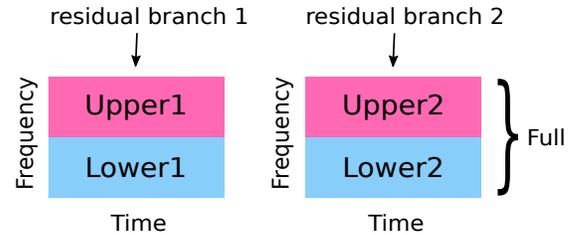}
  \caption{An illustration for the sub-band definitions}
\label{fig:shakesubband}
\end{figure}

Before we formally define the proposed models, we introduce the definition of sub-bands first. 
Fig. \ref{fig:shakesubband} depicts the definition for sub-bands in a $3$-branch residual block. 
Here we slightly abuse the notations of frequency and time because after two convolutional layers these axes are not exactly the same as those of input to the branches; 
however, since convolution is a local operation they still hold the corresponding spectral and temporal nature. 
At the output of each branch, we define the high-frequency half to be the upper sub-band while the low-frequency half to be the lower sub-band. 
We take the middle point on the spectral axis to be the border line for simplicity. 
The entire output is called the full band.

Having defined these concepts, we denote $\mathbf{X}$ the input to a residual block, $\mathbf{X}^i$ the full band from the $i$-th branch, $\mathbf{X}^i_u$ the upper sub-band from the $i$-th branch and $\mathbf{X}^i_l$ the lower sub-band from the $i$-th branch. Naturally, the relationship between them is given by $\mathbf{X}^i = \left[\mathbf{X}^i_u | \mathbf{X}^i_l\right]$. 
We also denote $\mathbf{Y}$ the output of a Shake-Shake regularized residual block.

To study the effectiveness of sub-band shaking for speech emotion recognition, we propose the following models for benchmarking:
\begin{enumerate}
\item Shake the full band (\textbf{Full})
\begin{equation}
\mathbf{Y} = \mathbf{X} + \sum_{n=1}^{N} \text{ShakeShake}\left(\left\{\mathbf{X}^n\right\}_{n=1}^{N}\right).
\label{eq:shake_full}
\end{equation}
\item Shake both sub-bands but independently (\textbf{Both})
\begin{eqnarray}
\mathbf{Y} &=& \mathbf{X} + \left[ \mathbf{Y}_u |\mathbf{Y}_l\right],
\label{eq:shake_both}\\
\mathbf{Y}_u &=& \sum_{n=1}^{N} \text{ShakeShake}\left(\left\{\mathbf{X}^n_u\right\}_{n=1}^{N}\right), \nonumber\\
\mathbf{Y}_l &=& \sum_{n=1}^N\text{ShakeShake} \left( \left\{ \mathbf{X}^n_l\right\}_{n=1}^{N}\right). \nonumber
\end{eqnarray}
\end{enumerate}

\section{Batch Normalized Shaking}
\label{sec:bn_shaking}
\begin{figure*}[htp]
  \centering
  \includegraphics[width=\linewidth, height=1.5in]{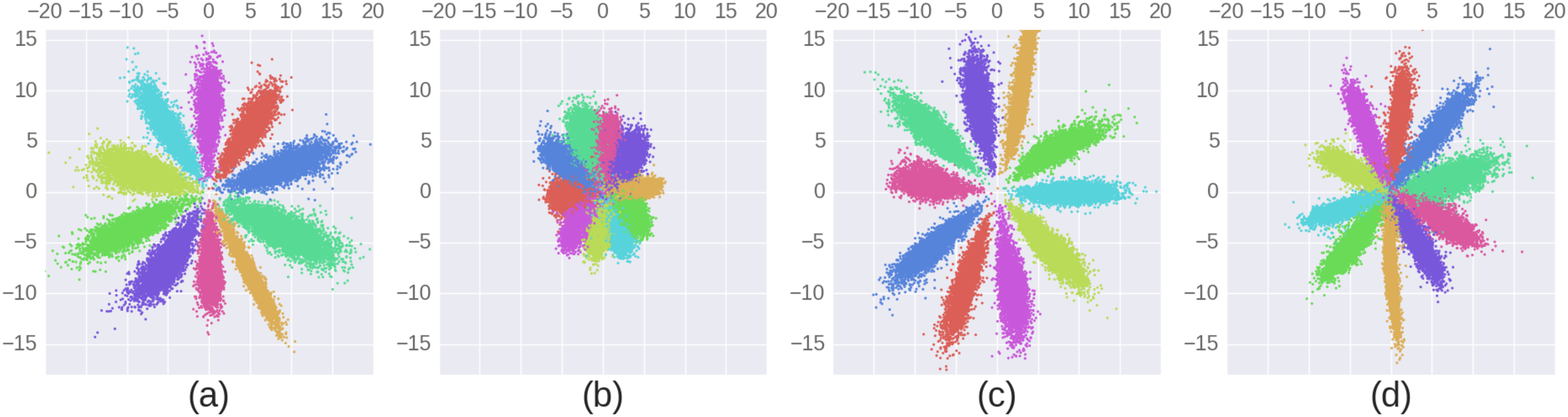}\\
  \includegraphics[width=\linewidth, height=1.5in]{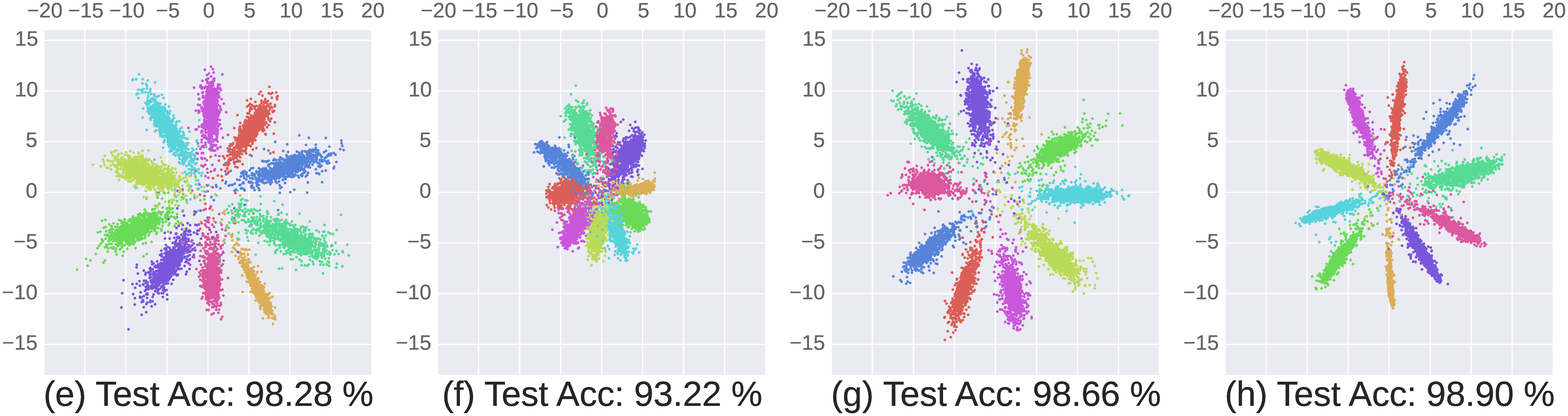}
  \caption{MNIST embeddings based on different layouts of residual blocks.
  We set the feature dimension entering into the output layer to be two and train them in an end-to-end fashion. 
  The top and bottom rows depict embeddings of the training samples extracted in the train mode (i.e. $\alpha\in[0,1])$ without updating parameters, and testing samples extracted in the eval mode ($\alpha=0.5$), respectively.
  (a,e) fully pre-activation (Fig. \ref{fig:resnetconfig}(c)) without shaking 
  (b,f) fully pre-activation (Fig. \ref{fig:resnetconfig}(c)) with shaking 
  (c,g) fully pre-activation + BN (Fig. \ref{fig:resnetconfig}(d)) without shaking
  (d,h) fully pre-activation + BN (Fig. \ref{fig:resnetconfig}(d)) with shaking}
  \label{fig:mnist_embed}
\end{figure*}

It has been shown that without batch normalization in a residual block, the shaking operation could easily cause training to diverge \cite{gastaldi2017}. 
Even with various methods to reduce the strength of shaking such as to limit the range of values $\alpha$ can take on or to decrease the number of residual blocks and hence the number of shaking layers, applying shaking regularization without batch normalization leads to a much inferior model, compared to a unregularized ResNeXt model with batch normalization.

\paragraph{\textbf{\textit{Batch Normalization for Symmetrically Distributed Representations}}}
From the perspective of discriminative feature learning, one may assume that without batch normalization, feature vectors from different classes may lie close to each other in an uneven fashion.
In this situation, the appropriate strength of shaking may be limited by the margin of two closest classes as in the reported experiments in \cite{gastaldi2017}.
In fact, a recent study \cite{batchnorm_smoother} refuted the commonly believed view on the role of batch normalization in reducing the so-called internal covariate shift.
Instead, they pointed out batch normalization actually helps to significantly smooth the optimization landscape during training.
Intuitively, it is expectedly easier to explore the neighborhood of well-separated class center vectors, compared to a set of unevenly distributed class center vectors, since it is less likely to have overlapping embeddings from different classes in the former case.
Without batch normalization, any inappropriately large strength of shaking may lead to \textit{overshot} embeddings, cause two or more classes to overlap and most importantly result in a slower and inferior convergence. 
Empirically, we find it is the key factor to keep a distribution of dispersed representations that spans over the entire feature space in a symmetric fashion and extends away from the origin, even when shaking is applied.
Therefore, we hypothesize that the batch normalization (or CCL) layer right before shaking serves to disperse embeddings and therefore prevent them from overlapping each other under the influence of perturbation that is encouraged by shaking around class center vectors.

\paragraph{\textbf{\textit{Model-Based Representation Augmentation for Coverage of Between-Class Feature Space}}}
From the review in Section \ref{sec:related_work}, we have noticed that classification tasks based on the VRM principle and verification tasks ask for subtly different distributions of representations.
The success of Mixup in classification tasks advocates for minimizing or eliminating the gap between classes in feature space by augmented training samples.
By doing so, the chance of predicting samples lying outside of the coverage of the original or augmented training samples, which leaves room for attacks by adversarial samples \cite{Szegedy2014} or results in uncertain predictions, is reduced.
On the other hand, CCL, and the other discriminative feature learning algorithms for verification, could leave a larger margin between classes (for example, c.f. Fig. \ref{fig:mnist_embed} (a) and (c)).
We hypothesize that shaking following a batch normalization layer expands the coverage of training samples and effectively it eliminates the margin between classes in feature space to achieve a similar distribution as in Mixup.
One caveat is that shaking does not smooth labels during exploration, and therefore is more similar to a prior work of Mixup, where interpolation and extrapolation of the nearest neighbors of the same class in feature space is proposed to enhance generalization \cite{DeVries2017}.
However, shaking operates solely on a single sample at a time, unlike \cite{Mixup2018} and \cite{DeVries2017}.

\subsection{Embedding Learning on MNIST and CIFAR-10}
\label{subsec:mnist_cifar10}
To demonstrate the close interaction between shaking and batch normalization in representation learning, we present two ablation studies, one on the MNIST dataset and the other on the CIFAR-10 dataset.
For both sets of experiments, we employ the ordinary softmax loss to examine the effectiveness of batch normalization in representation learning when it is not coupled with the softmax function.

The first set aims to visualize the embeddings of handwritten digit images learned under the influence of batch normalization and shaking.
We employ a ResNeXt (20, $2\times 4d$) architecture, where the last residual block reduces the feature dimension to $2$ for the purpose of visualization.
Fig. \ref{fig:mnist_embed} depicts embeddings by four layouts of residual blocks, where the top and bottom rows correspond to embeddings of the training and testing samples, respectively. 
From left to right, the columns represent embeddings learned by models of fully pre-activation (Fig. \ref{fig:resnetconfig}(c) \textbf{PreAct}) without shaking, \textbf{PreAct} with shaking, fully pre-activation + BN (Fig. \ref{fig:resnetconfig}(d) \textbf{PreActBN}) without shaking and \textbf{PreActBN} with shaking, respectively.

The first column serves as the baseline in this set of experiments.
Immediately, we can observe a severe degradation in separability when applying shaking to \textbf{PreAct}, comparing Fig. \ref{fig:mnist_embed}(a,e) with Fig. \ref{fig:mnist_embed}(b,f).
Also notice that shaking without a directly preceding batch normalization could perturb or destroy the symmetric distribution, which is obvious when there is no shaking (the symmetry in Fig. \ref{fig:mnist_embed}(a,e)).
This is rather interesting as batch normalization still exists in \textbf{PreAct} residual block, only not directly connected to the shaking layer.
It seems the exploration encouraged by shaking around each class center has expanded its coverage but without a directly preceding batch normalization to maintain a good dispersion between classes, each class only expands to overlap with neighboring classes,
and the resulting distribution is heavily tiled.
Consequently, \textbf{PreAct} with shaking delivers a much inferior performance compared to \textbf{PreAct}.

The comparison between \textbf{PreAct} (Fig. \ref{fig:mnist_embed}(a,e)) and \textbf{PreActBN} (Fig. \ref{fig:mnist_embed}(c,g)), both without shaking, demonstrates the effectiveness of CCL in discriminative feature learning although it is not coupled with the loss function.
In Fig. \ref{fig:mnist_embed}(c), not only does it maintain a symmetric distribution of classes, but also it encourages each class to expand outward and to leave more margin between neighboring classes.
As a result, \textbf{PreActBN} without shaking is able to reach a better performance compared to the baseline.

Finally, \textbf{PreActBN} with shaking (Fig. \ref{fig:mnist_embed}(d,h)), on the contrary, does not lead to a larger margin between classes as \textbf{PreActBN} without shaking does.
On the other hand, it seems that the shaking has expanded the coverage of each class so that all of them are directly adjacent to each other with a minimal or zero margin.
We could also observe that, although batch normalization tries to maintain a symmetric distribution of feature vectors, some of the classes in Fig. \ref{fig:mnist_embed}(d,h) are slightly tilted around the outer most region.
However, the most salient difference is the distribution of testing samples, where each class becomes more compact.
The performance of \textbf{PreActBN} is therefore the highest among all of the models.

\begin{figure}[htp]
  \centering
  \includegraphics[width=1\linewidth]{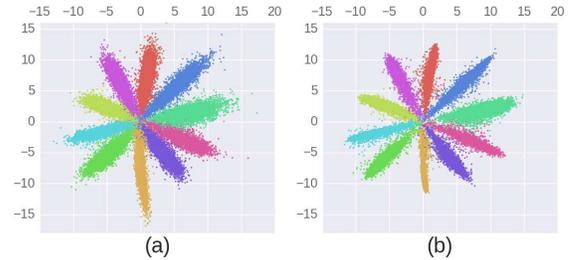}
  \caption{Embeddings of training samples extracted in the (a) train (b) eval mode from \textbf{PreActBN} with shaking}
  \label{fig:mnist_bnshake_train_eval}
\end{figure}

\begin{figure}[htp]
  \centering
  \includegraphics[width=1\linewidth]{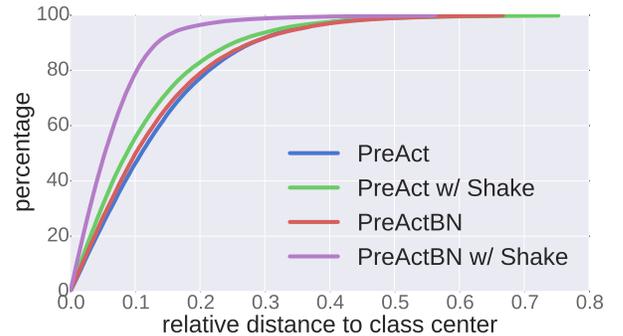}
  \caption{Percentage of class samples within in a relative distance to class center}
  \label{fig:relative_dist}
\end{figure}

Based on the VRM principle, the region close to boundaries between classes in feature space is covered by augmented training samples.
Since in supervised learning we assume testing samples are drawn from a similar or the same distribution as training samples, the majority of testing samples are mapped to embeddings close to the center of class, where most original training samples are mapped to, and thus the distribution seems more compact.
In Fig. \ref{fig:mnist_bnshake_train_eval}, from the comparison of training embeddings in the train and eval modes, it is visually clear that shaking is expanding the coverage of training embeddings.
To quantitatively show that this is the case, for each layout we calculate distances of the original training embeddings to their respective class center vectors in the eval mode and plot the percentage of class samples within distances relative to the largest distance in the class that is calculared in the train mode, in Fig. \ref{fig:relative_dist}.
It is clear that with shaking most of the original training embeddings are concentrated around the class center. For example, almost $100\%$ of original training embeddings lies within $0.3\times\text{largest distance}$ for \textbf{PreActBN} with shaking and just a small number of original training samples are lying close to the boundaries.

\begin{table}[h]
  \footnotesize
  \centering
  \begin{threeparttable}
  \begin{tabular}{cccr}
  \Xhline{6\arrayrulewidth}
  Model & Depth & Params & Error (\%)\\  
  \Xhline{4\arrayrulewidth}
  ResNeXt ($29$, $16\times 64d$) \cite{Xie2016} & $29$  & $68.1$M & $3.58$\\  
  \Xhline{\arrayrulewidth}
  Wide ResNet \cite{Zagoruyko2016WRN} & $28$ & $36.5$M & $3.80$\\
  \Xhline{4\arrayrulewidth}
  Shake-Shake ($26$, $2\times 96d$) \cite{gastaldi2017}  & $26$ & $26.2$M & $2.86$\\
  \Xhline{\arrayrulewidth}
  Shake-Shake ($26$, $2\times 64d$) \cite{gastaldi2017} & $26$ & $11.7$M & $2.98$\\
  \Xhline{\arrayrulewidth}
  ResNeXt ($26$, $2\times 64d$) \cite{gastaldi2017}  & $26$ & $11.7$M & $3.76$\\
  \Xhline{4\arrayrulewidth}
  PreActBN ($26$, $2\times 64d$)$^\dag$ & $26$ & $11.7$M & $^\ast2.95$\\
  \Xhline{\arrayrulewidth}
  BN-Shake ($26$, $2\times 64d$) & $26$ & $11.7$M & $^\ast3.65$\\
  \Xhline{\arrayrulewidth}
  PreAct ($26$, $2\times 64d$)$^\dag$ & $26$ & $11.7$M & $^\ast6.92$\\
  \Xhline{6\arrayrulewidth}
  \end{tabular}
  \begin{tablenotes}\footnotesize
  \item[$\ast$] average over three runs
  \item[$\dag$] with shaking
  \end{tablenotes}
  \end{threeparttable}
  \caption{Test error (\%) and model size on CIFAR-10}
  \label{tab:cifar10_exp}
\end{table}

The second set of experiments on CIFAR-10 is designed to measure the contribution of the last batch normalization in \textbf{PreActBN} with shaking.
In order to do so, we remove the first two batch normalization from the \textbf{PreActBN} residual block and rename the new one, the \textbf{BN-Shake} residual block (ReLU-Conv-ReLu-Conv-BN-Mul), assuming shaking is applied. 
Along with \textbf{PreActBN} and \textbf{PreAct}, by presenting \textbf{BN-Shake}, all of them with shaking,  we are able to quantitatively demonstrate the crucial location of batch normalization in a shaking regularized architecture.

We modify the open-sourced Torch-based Shake-Shake implementation\footnote{https://github.com/xgastaldi/shake-shake.} that was released with \cite{gastaldi2017} to  build these three architectures.
All of the rest of parameters such as the cosine learning rate scheduling and the number of epochs remain unchanged. 
Only the part that involves residual block definition is modified to serve our need.
We run each experiment for three times to obtain a robust estimate of the performance using different random seeds.
Table \ref{tab:cifar10_exp} presents the results of our experiments as well as the quoted performances on CIFAR-10 from \cite{gastaldi2017}.

Note that Shake-Shake ResNeXt in \cite{gastaldi2017} is based on the ReLu-only pre-activation residual block (Fig. \ref{fig:resnetconfig}(b) \textbf{RPreAct}) and is thus different from \textbf{PreAct} we have here.
Although the ResNeXt-$26$ $2\times64d$ is based on the \textbf{RPreAct} structure, with a shallow depth of $26$, it should be comparable to one that is based on the \textbf{PreAct} when no shaking is applied \cite{He_2016_ECCV}.
Therefore, we also take it as the baseline for the pre-activation layout.

The performance of \textbf{PreActBN} with shaking ($2.95\%$, mean of $2.89\%$, $3.00\%$ and $2.95\%$) is comparable to the reported performance of \textbf{RPreAct} with shaking ($2.98\%$), where both of them have a directly preceding batch normalization layer before shaking.
As expected, the performance of \textbf{PreAct} with shaking ($6.92\%$, mean of $6.76\%$, $6.82\%$ and $7.19\%$) is much worse than every model in Table \ref{tab:cifar10_exp}, including the baseline ResNeXt-$26$ $2\times64d$.

On the other hand, the result of \textbf{BN-Shake} ($3.65\%$, mean of $3.56\%$, $3.76\%$ and $3.62\%$) is rather positive.
With only one batch normalization layer, it outperforms not only \textbf{PreAct} with shaking but also the baseline ResNeXt-$26$ $2\times64d$.
This finding highlights the fact that the directly preceding batch normalization plays a crucial role in keeping a good dispersion of intermediate representations when shaking is applied to explore unseen feature space, while the dispersing effect of any other batch normalization that is separated by convolutional layers from the shaking layer is reduced  or only auxiliary.

Based on these two sets of experiments, it is safe to state that in the close interaction between batch normalization and shaking, batch normalization is mainly responsible for keeping a dispersed symmetric distribution of intermediate representations from perturbation by shaking, while the shaking mechanism expands the coverage of training samples to eliminate the margin between classes by promoting stochastic convex combinations of model branches, similar to \cite{Mixup2018} and \cite{DeVries2017}. 

For speech emotion recognition, we conduct experiments with $3$-branch ResNeXt based on the original (Fig. \ref{fig:resnetconfig}(a) \textbf{PostAct}), \textbf{RPreAct}, \textbf{PreActBN} and \textbf{PreAct} layouts with and without shaking to benchmark these architectures. 

\subsection{Relation with Batch Normalized Recurrent Neural Networks}
\label{subsec:bnrecurrent}
So far, we have presented arguments and analyses of experiments to clarify the interplay of batch normalization and shaking, and drawn a connection between each of them to discriminative feature learning and VRM, respectively. 

In our previous work on sub-band shaking \cite{Huang2018a} and stochastic Shake-Shake regularization \cite{Huang2018b} in affective computing, we found that shaking regularization contributes more to constraining the training process, measured in terms of the generalization gap between the accuracies of training and validation, than improving generalization when using \textbf{PreAct} with shaking, which is understandable based on previous analyses.
However, when employing \textbf{PreActBN} for speech emotion recognition, we further observe another interesting behavior of the resulting models.
Compared to \textbf{PreAct}, although the validation performance is improved thanks to the addition of batch normalization, the gap between training and validation accuracies also drastically increased, which suggests a significantly amount of increased over-fitting.

\begin{table}[ht]
  \footnotesize
  \centering
  \begin{threeparttable}
  \begin{tabular}{@{} lccr}
  \Xhline{4\arrayrulewidth}
  Architecture & Shake & Valid UA (\%) & Gap (\%)\\
  \Xhline{4\arrayrulewidth} 
  \multirow{ 2}{*}{PreAct}   & \xmark     & $61.342$ & $7.485$ \\ 
                             & \checkmark & $62.989$ & $\mathbf{-1.128}$ \\
  \Xhline{\arrayrulewidth}
  \multirow{ 2}{*}{PreActBN} & \xmark     & $63.407$ & $7.410$ \\ 
                             & \checkmark & $66.194$ & $\mathbf{8.348}$ \\
  \Xhline{4\arrayrulewidth}
  \end{tabular}
  \begin{tablenotes}\footnotesize
  \item[*] average over three runs
  \end{tablenotes}
  \end{threeparttable}
  \caption{Performances of \textbf{PreAct} and \textbf{PreActBN} with and without shaking for speech emotion recognition}
  \label{tab:preact_preactbn_shake_noshake}
\end{table}

For example, Table \ref{tab:preact_preactbn_shake_noshake} gives a comparison between \textbf{PreAct} and \textbf{PreActBN} with and without shaking.
With shaking, \textbf{PreActBN} achieves an improvement of $3.205\%$ on the un-weighted accuracy (UA) at the cost of an increment of $9.476\%$ on the generalization gap (Gap) between training and validation UAs from \textbf{PreAct}, simply because of the addition of batch normalization.
On the other hand, it is $2.787\%$ increment on the UA and $0.938\%$ increment on the Gap, respectively, for \textbf{PreActBN} due to the addition of shaking.
Apparently, the addition of batch normalization triggers something that causes this significant change in the generalization gap when shaking is applied.

We suspect that this behavior partially resembles the reported situation in application of batch normalization to recurrent neural networks \cite{Laurent2016,bnrecurrent2017},
where Laurent et al. \cite{Laurent2016} attempted to apply batch normalization to the recurrent formulation of temporal modeling and concluded that it leads to faster convergence but more over-fitting. 
To address this issue, Cooijmans et al. \cite{bnrecurrent2017} attributed the difficulties with recurrent batch normalization to gradient vanishing due to improper initialization of the batch normalization parameters, and proposed to initialize the standard deviation parameter $\gamma$ in Eq. (\ref{eq:bn}) to a small value such as $0.1$, contrary to the common practice of unit initialization.
With a proper initialization, they demonstrated successful applications of batch normalized Long Short Term Memory (LSTM) networks that converge faster and generalize better.

In fact, the formulations are awfully similar in batch normalized residual blocks and in batch normalized recurrent neural networks, where both of them consist of \textit{a batch normalized sum of multiple batch normalized streams and so on so forth}. 
For example, in \textbf{PreActBN} with shaking the output of the $l$-th block is as follows:
\begin{equation}
  \mathbf{X}_{l+1} = \alpha\text{BN}(f_l^1(\mathbf{X}_l)) + (1-\alpha)\text{BN}(f_l^2(\mathbf{X}_l)) + \mathbf{X}_l,
\end{equation}
and in the $(l+1)$-th block, 
\begin{eqnarray}
  f^i_{l+1}(\mathbf{X}_{l+1})
  &=&
  \text{Conv}(\text{ReLU}(\text{BN}(\text{Conv}(\mathbf{Y}^i)))), \\
  \mathbf{Y}^i &=& \text{ReLU}(\text{BN}(\mathbf{X}_{l+1})).
\end{eqnarray}

On the other hand, the batch normalized LSTM is given by the following equations \cite{bnrecurrent2017}:

\begin{eqnarray}
  \begin{pmatrix}
    \mathbf{f}_t \\
    \mathbf{i}_t \\
    \mathbf{o}_t \\
    \mathbf{g}_t 
  \end{pmatrix}
  &=&
  \text{BN}(\mathbf{W}_h\mathbf{h}_{t-1}) + \text{BN}(\mathbf{W}_x\mathbf{x}_t) + \mathbf{b}, \\
  \mathbf{c}_t
  &=&
  \sigma(\mathbf{f}_t) \odot \mathbf{c}_{t-1} + \sigma(\mathbf{i}_t) \odot \tanh(\mathbf{g}_t), \\
  \mathbf{h}_t
  &=&
  \sigma(\mathbf{o}_t) \odot \tanh(\text{BN}(\mathbf{c}_t)),
\end{eqnarray}
where $\mathbf{x}_t$ is the input, $\mathbf{W}_x$ and $\mathbf{W}_h$ are the input and recurrent weights, $\mathbf{f}_t$, $\mathbf{i}_t$, $\mathbf{o}_t$ and $\mathbf{g}_t$ are forget, input, output and gates and the cell candidate, $\mathbf{c}_t$ and $\mathbf{h}_t$ are the cell and output vectors, $\odot$ and $\sigma$ are the Hadamard product and the logistic sigmoid function, respectively.

In these two batch normalized formulations, we find $\mathbf{Y}^i$ and $\mathbf{h}_t$ rather similar to each other, up to only some component-wise scaling and clipping by activations.
In addition to benchmarking different layouts of residual blocks, we also present speech emotion recognition experiments to investigate the convergence behavior of shaking regularized ResNeXt networks with the batch normalization parameter $\gamma$ initialized to different values.

\section{Database and Experiments for Speech Emotion Recognition}
\label{sec:dataexps}

We use six publicly available emotion corpora to demonstrate the effectiveness of the proposed models, including the eNTERFACE'05 Audio-Visual Emotion Database \cite{enterface05}, the Ryerson Audio-Visual Database of Emotional Speech and Song (RAVDESS) \cite{ravdess}, the Interactive Emotional Dyadic Motion Capture (IEMOCAP) database \cite{Busso2008}, the Berlin Database of Emotional Speech (Emo-DB) \cite{BurkhardtPRSW05}, the EMOVO Corpus \cite{emovo} and the Surrey Audio-Visual Expressed Emotion (SAVEE) \cite{savee}. 
Some of these corpora are multi-modal in which speech, facial expression and text all convey a certain degree of affective information. 
However, in this paper we solely focus on the acoustic modality for experiments.

We formulate the experimental task into a sequence classification of $4$ classes, including joy, anger, sadness and fear. 
We perform sub-utterance sampling \cite{Keren16a} by dividing long utterances into several short segments of $6.4$-second long with the same label to limit the length of the longest utterance; 
for example, a $10$-second long angry utterance is replaced by two angry segments corresponding to the first $6.4$ and the last $6.4$ seconds of the original utterance with some overlapping part. 
In this way, we also slightly benefit from data augmentation. 
As a result, we obtain $6803$ emotional utterances from the aggregated corpora. 
Table \ref{tab:datasets} summarizes the information about these six corpora. 

\begin{table}[ht]
  \footnotesize
  \centering
  \begin{tabular}{@{} lccccc}
  \Xhline{4\arrayrulewidth}
  Corpus & No. & \multicolumn{4}{c}{No. Utterances}\\
  \cline{3-6}
  & Actors & Joy & Anger & Sadness & Fear \\
  \Xhline{2\arrayrulewidth} 
  eNTERFACE         & $42$ & $207$ & $210$ & $210$ & $210$ \\ 
  RAVDESS           & $24$ & $376$ & $376$ & $376$ & $376$\\
  IEMOCAP           & $10$ & $720$ & $1355$ & $1478$ & $0$ \\
  Emo-DB            & $10$ & $71$  & $127$  & $66$ &  $69$ \\
  EMOVO              & $6$  & $84$ & $84$ & $84$ & $84$\\
  SAVEE              & $4$  & $60$ & $60$ & $60$ & $60$\\
  \Xhline{2\arrayrulewidth} 
  Total        & $96$ & $1518$ & $2212$ & $2274$ & $799$\\
  \Xhline{4\arrayrulewidth}
  \end{tabular}
  \caption{An overview of these selected corpora, including the number of actors and the distribution of utterances in the emotional classes} 
  \label{tab:datasets}
\end{table}

\begin{table}[ht]
  \footnotesize
  \centering
  \begin{tabular}{lccccc}
  \Xhline{4\arrayrulewidth}
  Corpus & \multicolumn{4}{c}{Actor Set Partition} \\
  \cline{2-5}
  & 1 &  2 & 3 & 4 \\
  \Xhline{2\arrayrulewidth}
  eNTERFACE   & $2$F, $8$M & $2$F, $9$M & $3$F, $8$M & $2$F, $8$M\\ 
  RAVDESS     & $3$F, $3$M & $3$F, $3$M & $3$F, $3$M & $3$F, $3$M\\
  IEMOCAP     & $1$F, $2$M & $1$F, $1$M & $1$F, $1$M & $2$F, $1$M\\
  Emo-DB      & $1$F, $2$M & $1$F, $1$M & $1$F, $1$M & $2$F, $1$M\\
  EMOVO       & $1$F, $0$M & $1$F, $1$M & $1$F, $1$M & $0$F, $1$M\\
  SAVEE       & $0$F, $1$M & $0$F, $1$M & $0$F, $1$M & $0$F, $1$M\\
  \Xhline{2\arrayrulewidth} 
  Total       & $8$F, $16$M & $8$F, $16$M & $9$F, $15$M & $9$F, $15$M\\
  \Xhline{4\arrayrulewidth}
  \end{tabular}
  \caption{F: female, M: male. The gender and corpus distributions in each actor set partition of the cross validation} 
  \label{tab:cross-validation}
\end{table}

\begin{table}[h]
  \footnotesize
  \centering
  \begin{tabular}{cccc}
  \Xhline{4\arrayrulewidth}
  Layer Name & Structure & Stride & No. Params \\
  \Xhline{4\arrayrulewidth}
  prelim-conv & 
  $\begin{bmatrix}
  2\times16,  & 4  \\
  \end{bmatrix} \times 1$ & $[1,1]$ & $132$ \\
  \Xhline{\arrayrulewidth}
  res-8 & 
  $\begin{bmatrix}
  2\times16,  & 8  \\
  2\times16,  & 8  
  \end{bmatrix} \times 3$ & $[1,1]$ & $22.84$K\\
  \Xhline{\arrayrulewidth}
  res-16 & 
  $\begin{bmatrix}
  2\times16,  & 16  \\
  2\times16,  & 16  
  \end{bmatrix} \times 1$ & $[1,1]$ & $24.88$K\\
  \Xhline{\arrayrulewidth}
  res-32 & 
  $\begin{bmatrix}
  2\times16,  & 32  \\
  2\times16,  & 32  
  \end{bmatrix} \times 1$ & $[1,1]$ & $99.17$K\\
  \Xhline{\arrayrulewidth}
  average & - & - & - \\
  \Xhline{\arrayrulewidth}
  affine & $256\times 4$ & - & $1024$\\
  \Xhline{4\arrayrulewidth}
  total & -  & - & $148$K\\
  \Xhline{4\arrayrulewidth}
  \end{tabular}
  \caption{ResNeXt ($12, 2\times 8d$) network architecture }
  \label{tab:arch_resnet}
\end{table}

For the evaluation, we adopt a $4$-fold cross validation strategy. 
To begin with, we split the actor set into $4$ partitions. 
In addition, we impose extra constraints to make sure that each partition is as gender and corpus uniform as possible. 
For example, each actor set partition is randomly distributed with $2$-$3$ female actors and $8$-$9$ male actors from the eNTERFACE'05 corpus. 
More details are provided in Table \ref{tab:cross-validation}. 
By partitioning the actor set, it becomes easier to maintain speaker independence between training and validation throughout all of the experiments.   

We extract the spectrograms of each utterance with a $25$ms window for every $10$ms using the Kaldi \cite{Povey11thekaldi} library. 
Cepstral mean and variance normalization is then applied on the spectrogram frames per utterance. 
To equip each frame with a certain context, we splice it with $10$ frames on the left and $5$ frames on the right. 
Therefore, a resulting spliced frame has a resolution of $16\times257$. 
Since emotion involves a longer-term mental state transition, we further down-sample the frame rate by a factor of $8$ to simplify and expedite the training process. 

We build an architecture of ResNeXt ($12, 2\times 8d$) that consists of only $3$ residual stages, res-8, res-16 and res-32, in which the filter size is fixed to $2\times 16$. 
There are $3$ residual blocks in res-8, and one residual block in res-16 and res-32. 
Before the affine transformation, we add a dropout layer with probability of $0.5$.
Table \ref{tab:arch_resnet} contains details of the architecture.
For each utterance, a simple mean pooling is taken at the output of the final residual block to form an utterance representation before it is mapped by the affine layer. 
We avoid explicit temporal modeling layers such as a long short-term memory recurrent network because our focus is to investigate the effectiveness by shaking the ResNeXt. 
Note that a shaking layer has no parameter to learn and hence the model size in this work stays almost constant and only changes slightly in the situation when the number of batch normalization layers differs, since batch normalization has a set of trainable scaling and shift parameters.

We implement the shaking layer as well as the entire network architecture using the PyTorch \cite{pytorch} library. 
Only the Shake-Shake combination \cite{gastaldi2017} is used and shaking is applied independently per frame. 
We may also for simplicity refer to the Shake-Shake regularization as shaking regularization.
Due to class imbalance in the aggregated corpora, the objective function for training is the weighted cross-entropy, where the class weight is inversely proportional to the class size. 
The models are learned using the Adam optimizer \cite{KingmaB14} with an initial learning rate of $0.001$ and the training is carried out on an NVIDIA Tesla P100 GPU. 
We use a mini-batch of $8$ utterances across all model training and let each experiment run for $1000$ epochs, and for three runs with different random seeds to obtain a robust estimate of the performance.

For sub-band shaking experiments, we initially employ the fully pre-activation layout (Fig. \ref{fig:resnetconfig}(c)). \textbf{PreAct} denotes the model without shaking, while \textbf{PreAct-Full} and \textbf{PreAct-Both} correspond to the models with the \textbf{Full} and \textbf{Both} shaking (Eq. \ref{eq:shake_full}, \ref{eq:shake_both}), respectively. 
We only consider two sub-bands, defined by the middle point on the spectral axis, to present our point.
However, it does not rule out the feasibility of experiments with more sub-bands. 
In fact, in the multi-stream framework for speech recognition, more sub-bands were considered to generate more combinations and diversity.

We conduct experiments to benchmark different layouts, including \textbf{PostAct}  (Fig. \ref{fig:resnetconfig}(a)), \textbf{RPreAct} (Fig. \ref{fig:resnetconfig}(b)) and \textbf{PreActBN} (Fig. \ref{fig:resnetconfig}(d)) with and without (\textbf{Full}) shaking.  
Furthermore, we choose three different initialization values of $\gamma$ in batch normalization and apply them to each of \textbf{PostAct}, \textbf{RPreAct} and \textbf{PreActBN} layouts to investigate if batch normalized ResNeXt networks also behave as batch normalized recurrent neural networks in terms of faster convergence and better generalization, in addition to the resemblance in mathematical formulations.
In this set of experiments, we use the affine transformation in batch normalization despite CCL recommends not to use it, except in the last batch normalization before the output layer, where we only standardize the input and do not rescale it.

Finally, we revisit sub-band shaking with the \textbf{PreActBN} layout and with different initialization values of $\gamma$ to present a complete comparison between shaking on the full band and on the sub-bands independently. 
With this set of experiments, we may also examine the effectiveness of trade-off between the un-weighted accuracy (UA) and the generalization gap (Gap) between training and validation UAs when batch normalization is applied right before shaking.

\section{Experimental Results}
\label{sec:exp_results}
\subsection{Sub-band Shaking}
\label{subsec:exp_subband_shaking}
The result of sub-band shaking is presented in Table \ref{tab:ua_gap_subband_shake}.
It is clear that both \textbf{PreAct-Full} and \textbf{PreAct-Both} are able to improve from \textbf{PreAct}, unlike the experiments on MNIST and CIFAR-10 datasets.
For both shaking regularized models, however, their improvements on the UA is comparably smaller than their reduction on the generalization gap, which suggests the models are learning a harder problem in the presence of shaking.
It could also be the difference in the nature of tasks or in the number of classes that \textbf{PreAct} with shaking is improving instead of degrading.

\begin{table}[ht]
  \footnotesize
  \centering
  \begin{threeparttable}
  \begin{tabular}{crr}
  \Xhline{6\arrayrulewidth}
  Model & Valid UA (\%) & Gap (\%)  \\
  \Xhline{4\arrayrulewidth}
  PreAct & $61.342$ & $7.485$ \\
  \Xhline{\arrayrulewidth}
  PreAct-Full & $^\dag 62.989$  & $^\dag -1.128$ \\
  \Xhline{\arrayrulewidth}
  PreAct-Both & $^\dag 64.973$ & $^\dag 1.791$ \\
  \Xhline{6\arrayrulewidth}
  \end{tabular}
  \begin{tablenotes}\footnotesize
  \item[*] average over three runs
  \item[$\dagger$] significantly ($p$$<$$0.05$) outperforms \textbf{PreAct}
  \end{tablenotes}
  \end{threeparttable}
  \caption{Performances of \textbf{PreAct}, \textbf{PreAct-Full} and \textbf{PreAct-Both}}
  \label{tab:ua_gap_subband_shake}
\end{table}

Furthermore, we conduct statistical hypothesis testing to examine the significance of the observed improvements.
The resulting p-values from one-sided paired t-test indicates that the improvements of \textbf{PreAct-Full} and \textbf{PreAct-Both} from \textbf{PreAct} are statistical significant, in terms of the UA and the Gap.
In addition, compared to the improvements of \textbf{PreAct-Full}, \textbf{PreAct-Both} has reached a higher UA but also a larger gap. 
In fact, the improvement of \textbf{PreAct-Both} from \textbf{PreAct-Full} is also significant in terms of UA. 
On the other hand, \textbf{PreAct-Full} has a significant reduction on the Gap, compared to the reduction of \textbf{PreAct-Both} on the Gap. 
More details are presented in Table \ref{tab:p_value_subband_shake}.

\begin{table}[ht]
  \footnotesize
  \centering
  \begin{threeparttable}
  \begin{tabular}{crr}
  \Xhline{6\arrayrulewidth}
  \multicolumn{1}{c}{Model Pair} & 
  \multicolumn{1}{c}{Valid UA} & 
  \multicolumn{1}{c}{Gap}\\
  \Xhline{4\arrayrulewidth}
  PreAct vs PreAct-Full & $1.44\times10^{-2}$ & $2.7\times10^{-4}$\\
  \Xhline{\arrayrulewidth}
  PreAct vs PreAct-Both & $1.22\times10^{-5}$ & $3.0\times10^{-3}$ \\
  \Xhline{\arrayrulewidth}
  PreAct-Full vs PreAct-Both & $2.60\times10^{-3}$ & $9.9\times10^{-1}$ \\
  \Xhline{6\arrayrulewidth}
  \end{tabular}
  \end{threeparttable}
  \caption{P-values resulted from one-sided paired t-test with df=$4\times3-1=11$}
  \label{tab:p_value_subband_shake}
\end{table}

Based on these observations, we can think of sub-band shaking as a method to relax the strength of shaking. 
This way, one may trade off between the amount of improvement on the UA and the amount of reduction on the Gap.
Our previous work on stochastic Shake-Shake regularization \cite{Huang2018b} also investigated another method to trade off between these two performance metrics by randomly turning off shaking.
Since sub-band shaking and stochastic Shake-Shake behave similarly and the latter do not contribute directly into the focus of this work, we choose to present only sub-band shaking here.

\subsection{Experiments with Different Layouts}
\label{subsec:exp_shake_noshake}
The results of benchmarking the layouts are presented in Table \ref{tab:preact_preactbn_postact_rpreact_shake_noshake}.
First of all, we notice that even without shaking, all of \textbf{PostAct}, \textbf{RPreAct} and \textbf{PreActBN} have achieved higher UAs with significance (denoted by $\star$), compared to that of \textbf{PreAct}, while the UA of the first three models are comparable, i.e. no statistical significance in the differences.
These improvements on the UA could be attributed to discriminative feature learning, i.e. CCL, since we additionally apply batch normalization without rescaling before the final affine layer in the first three models.
We deliberately employ CCL in these three models to set up a set of competitive baselines.

Next, we look at the layouts with shaking at $\gamma_0=1.0$.
We immediately observe that even with the common practice of unit initialization for $\gamma$, all of these three layouts give further improvements on the UA.
However, none of them is able to reduce the generalization gap with statistical significance.
The closest one is \textbf{RPreAct} with shaking when $\gamma_0=1.0$ and the p-value for the testing is $0.088$ due to a high variation of the generalization gap from each fold.
With the common practice of unit initialization, \textbf{PreActBN} converges to the most accurate model with the UA as high as $66.194\%$, but it also results in the largest generalization gap.

\begin{table}[ht]
  \footnotesize
  \centering
  \begin{threeparttable}
  \begin{tabular}{@{} lccrr}
  \Xhline{4\arrayrulewidth}
  Architecture & Shake & $\gamma_0$ & Valid UA (\%) & Gap (\%)\\
  \Xhline{4\arrayrulewidth} 
  \multirow{ 2}{*}{PreAct}   & \xmark     &--&     $61.342$ & $7.485$ \\  
                             & \checkmark &--&     $^\dag 62.989$ & $^\dag -1.128$ \\
  \Xhline{\arrayrulewidth}
  \multirow{ 5}{*}{PostAct} & \xmark      &--&     $^\star62.782$ & $7.822$ \\  
                            & \checkmark  &$1.00$& $^\dag 65.536$ & $7.657$ \\
                            & \checkmark  &$0.20$& $^\dag 65.490$ & $5.431$ \\
                            & \checkmark  &$0.10$& $^\dag 65.483$ & $7.255$ \\
                            & \checkmark  &$0.05$& $^\dag 65.512$ & $7.205$ \\
  \Xhline{\arrayrulewidth}
  \multirow{ 5}{*}{RPreAct} & \xmark      &--&     $^\star63.939$ & $8.517$ \\  
                            & \checkmark  &$1.00$& $^\dag 65.859$ & $7.063$ \\
                            & \checkmark  &$0.20$& $^\dag 65.821$ & $8.305$ \\
                            & \checkmark  &$0.10$& $^\dag 64.899$ & $7.958$ \\
                            & \checkmark  &$0.05$& $^\dag 65.052$ & $^\dag 5.821$ \\
  \Xhline{\arrayrulewidth}
  \multirow{ 5}{*}{PreActBN} & \xmark     &--&     $^\star63.407$ & $7.410$ \\  
                             & \checkmark &$1.00$& $^\dag 66.194$ & $8.348$ \\
                             & \checkmark &$0.20$& $^\dag 65.789$ & $^\ddagger 5.817$ \\
                             & \checkmark &$0.10$& $^\dag 66.097$ & $^\ddagger 6.040$ \\
                             & \checkmark &$0.05$& $^\dag 66.418$ & $^\dag$ $^\ddagger 3.416$\\
  \Xhline{4\arrayrulewidth}
  \end{tabular}
  \begin{tablenotes}\footnotesize
  \item[*] average over three runs
  \item[$\star$] significantly ($p$$<$$0.05$) outperforms \textbf{PreAct} without shaking 
  \item[$\dagger$] significantly ($p$$<$$0.05$) outperforms the same layout without shaking
  \item[$\ddagger$] significantly ($p$$<$$0.05$) outperforms the same layout at $\gamma_0=1.0$
  \end{tablenotes}
  \end{threeparttable}
  \caption{Performances of \textbf{PreAct}, \textbf{PostAct}, \textbf{RPreAct} and \textbf{PreActBN} with and without shaking for speech emotion recognition, where $\gamma_0$ is the initialization value of the standard deviation parameter $\gamma$ in batch normalization}
  \label{tab:preact_preactbn_postact_rpreact_shake_noshake}
\end{table}

In fact, we could see that for all values of $\gamma_0$ we choose, these three layouts all outperform their respective baseline models, the same layout without shaking, in terms of UA with significance (denoted by $\dag$), and all of them show a trend of generalization gap reduction only that the majority of them are unable to reduce the generalization gap with significance, except two of them.
Both of \textbf{RPreAct} and \textbf{PreActBN} outperform their baseline models simultaneously on the UA and on the Gap, achieving a higher accuracy as well as reduced over-fitting at the same time, when $\gamma$ is initialized as $0.05$.
\textbf{PreActBN} with shaking at $\gamma_0=0.05$ is also the best model, both highest on the UA and lowest on the Gap (except for \textbf{PreAct} with shaking, which is known to be difficult to learn).
In other words, the result of this model validates our hypothesis with statistical significance that batch normalized ResNeXt networks are locally in formulation similar to batch normalized recurrent neural networks, and both of them require a careful selection of initialization values for batch normalization parameters to avoid the aforementioned difficulties. 

For comparison between layouts, we find \textbf{PreActBN} with shaking at $\gamma_0=0.05$ also outperforms most of the settings in \textbf{PostAct} and \textbf{RPreAct} layouts with statistical significance, except for the reduction on the generalization gap by \textbf{PostAct} at $\gamma_0=0.2$, where \textbf{PreActBN}@0.05 is not better with significance.
We summarize the complete comparison results in Table \ref{tab:preactbn_rpreactbn_postactbn_cf}.

\begin{table}[ht]
  \footnotesize
  \centering
  \begin{threeparttable}
  \begin{tabular}{@{} lcrr}
  \Xhline{4\arrayrulewidth}
  Architecture Pair & $\gamma_0$ & Valid UA & Gap \\
  \Xhline{4\arrayrulewidth} 
  \multirow{ 5}{*}{PreActBN@0.05 vs PostAct}  &$1.00$& $ 2.37\times10^{-3}$ & $4.12\times10^{-2}$ \\
                                         &$0.20$& $ 2.96\times10^{-3}$ & $6.44\times10^{-2}$ \\
                                         &$0.10$& $ 1.71\times10^{-2}$ & $2.15\times10^{-2}$ \\
                                         &$0.05$& $ 2.21\times10^{-2}$ & $1.27\times10^{-2}$ \\
  \Xhline{\arrayrulewidth}
  \multirow{ 5}{*}{PreActBN@0.05 vs RPreAct}  &$1.00$& $ 1.29\times10^{-2}$ & $1.00\times10^{-2}$ \\
                                         &$0.20$& $ 2.85\times10^{-2}$ & $2.44\times10^{-3}$ \\
                                         &$0.10$& $ 5.07\times10^{-4}$ & $1.34\times10^{-2}$ \\
                                         &$0.05$& $ 4.14\times10^{-3}$ & $4.23\times10^{-2}$ \\
  \Xhline{4\arrayrulewidth}
  \end{tabular}
  \end{threeparttable}
  \caption{P-values from one-sided paired t-test between \textbf{PreActBN} at $\gamma_0=0.05$ and \textbf{PostAct} and \textbf{RPreAct} with various value of $\gamma_0$}
  \label{tab:preactbn_rpreactbn_postactbn_cf}
\end{table}

\begin{table}[ht]
  \footnotesize
  \centering
  \begin{threeparttable}
  \begin{tabular}{@{} lccrr}
  \Xhline{4\arrayrulewidth}
  Architecture & Shake & $\gamma_0$ & Valid UA (\%) & Gap (\%)\\
  \Xhline{4\arrayrulewidth}
  PreAct                     & \checkmark &$1.00$& $64.973$ &$1.791$ \\
  \Xhline{\arrayrulewidth}
  \multirow{ 5}{*}{PreActBN} & \checkmark &$1.00$& $65.544$ &$8.034$ \\
                             & \checkmark &$0.20$& $65.679$ &$^\ddagger4.408$ \\
                             & \checkmark &$0.10$& $66.170$ &$6.257$ \\
                             & \checkmark &$0.05$& $^\ddagger66.432$ &$^\ddagger5.539$ \\
  \Xhline{4\arrayrulewidth}
  \end{tabular}
  \begin{tablenotes}\footnotesize
  \item[*] average over three runs
  \item[$\ddagger$] significantly ($p$$<$$0.05$) outperforms the same layout at $\gamma_0=1.0$
  \end{tablenotes}
  \end{threeparttable}
  \caption{Performances of \textbf{PreAct} and \textbf{PreActBN} with sub-band shaking for speech emotion recognition, where $\gamma_0$ is the initialization value of the standard deviation parameter $\gamma$ in batch normalization}
  \label{tab:preactbn_shake_subband_init}
\end{table}

Finally, when comparing $\gamma_0$-initialized batch normalized networks (when $\gamma_0=0.05,0.1,0.2$) with their counterpart by unit initialization, we find that although both \textbf{RPreAct} and \textbf{PreActBN} show a tendency to reduce the Gap with a smaller $\gamma_0$, only \textbf{PreActBN} manages to outperform its unit-initialized counterpart (denoted by $\ddagger$).
It may suggest that the extra batch normalization between residual blocks in \textbf{PreActBN} makes it structurally more similar to batch normalized recurrent neural networks.
On the other hand, any two batch normalization layers in \textbf{PostAct} or \textbf{RPreAct} are always separated by one convolutional layer.

In Section \ref{subsec:exp_subband_shaking}, we have seen that sub-band shaking leads to an improvement on the UA but also enlarges the generalization gap.
However, instead of \textbf{PreActBN} we employed \textbf{PreAct} previously and did not experiment with different initialization values of $\gamma$.
To present a complete comparison between shaking on the full band and on the sub-bands independently, we conduct experiments on sub-band shaking with the \textbf{PreActBN} layout and with the chosen initialization values of $\gamma$.
The results are summarized in Table \ref{tab:preactbn_shake_subband_init}.
Again, comparing \textbf{PreAct} and \textbf{PreActBN} both at $\gamma_0=1.00$, we could easily observe a significant increment of the generalization gap along with the additional batch normalization. 
With a smaller initialization value, not only does \textbf{PreActBN} reduce the generalization gap but also improve the UA when $\gamma_0=0.05$ with statistical significance.
Yet, the resulting $5.539\%$ on the Gap is still significantly larger than $1.791\%$ on the Gap by the \textbf{PreAct} layout.

We further compare the performance of \textbf{PreActBN} with shaking on the full band ($66.418\%$/$3.416\%$) and on the sub-bands independently ($66.432\%$/$5.539\%$).
The p-values for these two settings are $0.51$ and $0.079$ for testing between the UAs and the Gaps.
In the end, shaking on the full band and on the sub-bands independently give comparable performances with shaking on the full band slightly better at reducing the generalization gap.

\section{Conclusion}
\label{sec:conclu}
We investigate the recently proposed Shake-Shake regularization and its variants for classification in general and speech emotion recognition in particular.
In order to explain the observed interaction between batch normalization and shaking regularization, we base our ablation analysis of batch normalization in the MNIST experiments on discriminative feature learning.
Our experiments show that the batch normalization right before shaking regularization is crucial in that it keeps a dispersed symmetric distribution of intermediate representations from being tilted by random perturbation due to shaking.
Without the batch normalization, shaking regularization could easily tilt the distribution of each classes and cause them to overlap each other.

In addition, we highlight the subtle difference in the requirement of embeddings for classification tasks and verification tasks, where according to the vicinal empirical minimization principle or the recent success of Mixup \cite{Mixup2018}, classification tasks should try to minimize the margin between classes in feature space, while verification tasks aim to find an embedding distribution that has a minimal intra-class variation but a maximal inter-class dispersion.
From this perspective, we find the embeddings by \textbf{PreActBN} with shaking are indeed distributed with small or zero margin between classes (Fig. \ref{fig:mnist_embed}(d)), while the embeddings by \textbf{PreActBN} without shaking are distributed with large margins (Fig. \ref{fig:mnist_embed}(c)).
Moreover, since the original embeddings are distributed close to the class center vectors, the distribution of testing embeddings, which is often assumed to be the same or similar distribution of training samples, becomes more compact and hence leads to a higher accuracy.
This finding provides a direct explanation based on the VRM principle for the ability of shaking regularization to help improve classification tasks.
Another set of experiments on CIFAR-10 further validates it is the direct concatenation of batch normalization and shaking that contributes the most to the improvement on classification accuracy, and other layers of batch normalization are auxiliary.

In addition, we find that the formulation of batch normalized residual blocks resemble batch normalized recurrent neural networks.
Moreover, both of these two architectures suffer from a reported issue of fast convergence but more over-fitting.
To reduce the observe increment of the generalization gap in our speech emotion recognition experiments, we properly initialize the $\gamma$ parameter with a smaller value.
The experimental results validate our hypothesis and give a significant reduction on the generalization gap while achieving the same or a higher improvement on the UA.

A final comparison between shaking on the full band and on the sub-bands independently shows that with the additional batch normalization in \textbf{PreActBN} and a proper initialization value of $\gamma$, the difference between these two kinds of shaking is minimized, only that shaking on the full band slightly better at reducing the generalization gap.

\bibliographystyle{IEEEtran}
\bibliography{mybib}

\end{document}